\documentclass[runningheads]{llncs}
\usepackage{graphicx}
\usepackage{url}
\usepackage{todonotes}
\usepackage[]{algorithm2e}
\usepackage[utf8]{inputenc}
\usepackage[T1]{fontenc}
 \usepackage{booktabs}

\begin{document}
\title{Generating Philosophical Statements using Interpolated Markov Models and Dynamic Templates}
\titlerunning{Generating Philosophical Statements}
\author{Thomas Winters\orcidID{0000-0001-7494-2453}}
\authorrunning{T. Winters}
\institute{KU Leuven, Belgium \\
\email{thomas.winters@cs.kuleuven.be}}
\maketitle
\begin{abstract}
    Automatically imitating input text is a common task in natural language generation, often used to create humorous results.
    Classic algorithms for learning to imitate text, e.g. simple Markov chains, usually have a trade-off between originality and syntactic correctness.
    We present two ways of automatically parodying philosophical statements from examples overcoming this issue, and show how these can work in interactive systems as well.
    The first algorithm uses interpolated Markov models with extensions to improve the quality of the generated texts.
    For the second algorithm, we propose dynamically extracting templates and filling these with new content.
    To illustrate these algorithms, we implemented TorfsBot, a Twitterbot imitating the witty, semi-philosophical tweets of professor Rik Torfs, the previous KU Leuven rector.
    We found that users preferred generative models that focused on local coherent sentences, rather than those mimicking the global structure of a philosophical statement.
    The proposed algorithms are thus valuable new tools for automatic parody as well as template learning systems.

\end{abstract}
\section{Introduction}

Imitating input text is a popular natural language generation task.
One way of achieving this is by manually writing an context-free grammar, which often results in a relatively small and predictable generative space due to the large effort it takes to create such a grammar.
Markov chains are a useful tool for learning to imitate input sequences automatically.
These usually work best for generating short texts, as they tend to derail quickly due to their limited memory.
A textual Markov chain learner calculates how often a certain word follows the $n$ previous words \cite{norris1997markov-chains}.
One consideration designers of such systems have is choosing this $n$: a high number enforces grammatically more sensible sentences, but more source text plagiarism, whereas a low number leads to more random but original output \cite{papadopoulos2014plagiarism}.
In this paper, we propose two potential solutions for this trade-off for generating short, philosophical statements, by developing an algorithm that generates more locally coherent text, and another algorithm that mimics the global structure of a philosophical statement .
We demonstrate our algorithms by training on the tweets and columns of the previous KU Leuven rector, Rik Torfs, and deploying it as a Twitterbot called TorfsBot.

\section{Method}

\subsection{Interpolated Markov Model}

A first solution to partially mitigate the classic trade-off between originality and syntactic correctness is by interpolating between several different Markov models.
That is, the algorithm trains multiple Markov models with a different order (e.g. with $k\in\{2,3,4\}$), and gives a weight to every Markov model (e.g. $w_2=1, w_3=200,w_4=400$).
When the interpolated Markov model chooses a next word, the count of every possible candidate is multiplied with the weight of the Markov model it originates from.
It then selects the next word from all weighted candidates using roulette wheel selection on the summed counts of all candidates.
In this step, additional heuristic functions $f$ could further influence the weights, e.g. preferring certain topical words, rhymes etc.
In summary, the model thus does not only look at the previous two words to know the probability of of the next word, but also at the previous three and previous four words, sums all probabilities of all possible words according to these other models, but gives a higher probability of picking a word predicted by a model looking at more previous words.

Thus, the new formula to calculate the probability of candidate words can be derived from the classic Markov model by adding some extensions.
The classic Markov model of order $k$, which looks back $k$ words to predict the probability of word $s_{i+1} \in$ vocabulary $V$ from previous words $s_0, s_1, \ldots, s_i$, uses the following formula \cite{papadopoulos2014plagiarism}:

\begin{equation}
p(s_{i+1} | s_{0}, s_{1}, \ldots, s_{i}) = p(s_{i+1} | s_{i-k}, s_{i-k+1}, \ldots, s_{i})
\end{equation}

with $p(s_{i+1} | s_{i-k}, s_{i-k+1}, \ldots, s_{i})$ the number of times $s_{i+1}$ occurred after $s_{i-k}$, $s_{i-k+1}$, $\ldots$, $s_{i}$ in the training corpus, divided by the number of times $s_{i-k}$, $s_{i-k+1}$, $\ldots$, $s_{i}$  followed by any word occurred.

We first extend this with a normalisation function $n$, increasing the variety while decreasing the size of the resulting automaton. This function maps the words to lowercase and removes any non-alphanumeric character.

\begin{equation}
p(s_{i+1} | s_{0}, s_{1}, \ldots, s_{i}) = p(s_{i+1} | n(s_{i-k}), n(s_{i-k+1}), \ldots, n(s_{i}))
\end{equation}

The following formula combines evidence from multiple such Markov models with maximum order $k$ by weighting every model using weight $w_j$, creating an interpolated Markov model:

\begin{equation}
p(s_{i+1} | s_{0}, s_{1}, \ldots, s_{i}) = \frac{\sum_{j=1}^k w_j \cdot p(s_{i+1} | n(s_{i-j}), n(s_{i-j+1}), \ldots, n(s_{i})) }{\sum_{j=1}^{k}w_j}
\end{equation}

These weights can then be further influenced by a function $f$, which increases the weight if the new word fulfills certain conditions, such as having close phonetic distance, contextually related to previously generated words, or combinations thereof:

\begin{equation}
\label{eq:weight-influence-markov}
p(s_{i+1} | s_{0}, s_{1}, \ldots, s_{i}) =
\frac{
\sum_{j=1}^k w_j \cdot p(s_{i+1} | n(s_{i-j}), \ldots, n(s_{i}))
\cdot
f(s_{i+1}, \ldots, s_1)
}
{
\sum_{v \in V}
\sum_{j=1}^k w_j \cdot p(v | n(s_{i-j}), \ldots, n(s_{i}))
\cdot
f(v, s_i, \ldots, s_1)
}
\end{equation}

The algorithm uses any start of a sentence in the training data as $s1$, $\ldots$, $s_{min_k(w_k \neq 0)}$ to kick start the generatibe process using Equation \ref{eq:weight-influence-markov}, until an ending token is obtained as $s_{i+1}$.
After generating a candidate, the algorithm performs several post-processing extensions for improving the quality of the final generated texts.

\subsubsection{Shortening}

Since philosophical statements and tweets tend to be better when short, the first post-processing step of the algorithm shortens generated text if it is longer than a specific threshold.
For our purposes, we set this threshold to the previous Twitter limit of 140 characters.
It achieves this shortening by removing sentences between the first and last sentence, and leaving some middle sentences if there is enough space left.
Due to the fact that the Markov chains enforce that both the beginning and the endings truly occur as beginnings and endings in the training data, they tend to feel as respectively introductions and conclusions.
This pattern of skipping intermediary sentences, and thus just going from introduction to conclusion, is a property also generally present in philosophical statements and tweets.
Another way the algorithm achieves snappy texts is by first generating multiple candidates and then preferring shorter ones by using roulette wheel selection using the inverse text length as the main factor of the weight.

\subsubsection{Punctuation fixing}

As a second post-processing step, the algorithm fixes the brackets and quotation marks by adding complementary brackets and quotation marks at the beginning or endings of clauses or sentences.
If no good position for the other bracket can be found, the bracket or quotation mark is deleted.
This is a necessary step due to the fact that Markov models are only able to keep track of the last several generated words.
Contrary to many neural networks text generators \cite{karpathy2015rnn,radford2019gpt2}, these Markov models are thus most of the time unaware what the current level of nested brackets is during generation.
As such, this post-processing step fixes several obvious mistakes in the generated texts.

\subsubsection{News insertion}
Since tweets are often bound to events happening when the tweet was written, using them as training data might result in referencing past dates, events and news.
To mitigate this, the third post-processing step is to link specifics in the generated text such to the present and to current events, as this makes a text feel more relevant and thus improves its quality \cite{winters2019clin}.
To achieve this, the algorithm searches for occurrences of full dates, months and years using regular expressions, and replaces them with a (parts of) a random date in the near future; we chose one to four days.
It also searches the text for named entities, as these are either tied to events from the time it was written, or serve as good candidates for adding references to current events.
It finds these by searching for title-cased words in unexpected spots, as this is a well-working proxy for named entity recognition in our target languages.
The system filters out named entities that occur more than a specific threshold in the training corpus, such that it does not replace names that are archetypal references for the author the system imitates.
The system then crawls the front page of a news website and selects a news article based on unigram similarity to the training corpus, where the score is calculated in a way similar to the scores of the naive Bayes algorithm.
It uses this article to replace the filtered names from the generated text by the named entities of the chosen news article, giving priority to the most frequent named entities in the article.
If the generated text contains any quotes that are longer than three words (and thus a proper quote, and not word emphasis), it also replaces the quote with a quote from the article if there is one.
All these replacements help the generated text feel more grounded in the present and current affairs.

\subsubsection{Originality check}

The final post-processing step is checking if the generated text is original enough compared to the training corpus.
While there are clever extensions to create a Markov automaton that checks for originality while generating \cite{papadopoulos2014plagiarism}, we opted for a post-processing step due to being less memory intensive  as well as the limited length of the generated texts.
This is achieved by normalising both by converting to lower case and removing non-alphanumeric characters, and checking if the normalised generated text occurs fully in, or as a part of an element of the normalised training corpus.
If there is overlap, it restarts the full process and generates a new text.

\subsection{Dynamic Templates}

Templates are often used for generating text, and are especially popular in computational humour \cite{binsted1994jape,manurung2008standup,winters2018automaticjokegeneration,winters2019goofer}.
A template is essentially a sentence with slots, where every slot is a variable that a data source can fill.
Context-free grammars allow designers to generate texts using specified templates, and have been used to create thousands of Twitterbots \cite{compton2017shared}, e.g. by using the popular Tracery tool \cite{compton2015tracery}.
However, creating such a grammar by hand is not only tedious, but also tends to caricaturize the target due to having a limited number of templates.

We propose solving the issue of creating templates by hand by dynamically extracting templates from base texts by knowing what types of content we want to insert, and identify good slots for replacing words by identifying the key words of a chosen sentence.
This way, the global structure of the statement is retained.
These base texts with yet-to-define slots is what we call ``dynamic templates''.

To generate a text using dynamic templates, the algorithm first receives a corpus of base texts $B$, a corpus of content text $C$ and a unigram model $U$ as input.
The algorithm then picks several consecutive lines from $C$ and uses their words as a proxy for finding a set of related context words.
It then creates a mapping from their part-of-speech tags to every present word.
For the dynamic template itself, it selects a random text from the base text corpus $B$ and analyses all the present part-of-speech tags.
The algorithm then proposes a list of possible replacements based on matching part-of-speech tags and the mapping to content text words.
The part-of-speech tags do not only contain information about the type, but also more specific information e.g. the tense of a verb.
This ensures that words that are replaced generally still have the correct syntactical relation to other words in the sentence.
Words that do not have matching part-of-speech tags present in the selected lines from $C$ are not replaced.
Some types of words, such as auxiliary verbs and other structural words, are also not allowed to be replaced due to breaking the grammatical correctness of the sentence.
The possible replacements are sorted by ascending frequency using the unigram model $U$, to prefer rare words, as a proxy for the key contextual words.
The algorithm must replace a certain minimum number of words, proportional to the length of the string, since we do not want short sentences with too common words without replacements, or long sentences with only one word being replaced.
The replacement process continues until words to be replaced are more common than a minimum percentile in $U$, modelling the word commonness distribution.
Similar to the interpolated Markov model model, it then also performs the news insertion step to make the generated texts more tied to current events.
The pseudocode for this algorithm can be found in Algorithm \ref{alg:dynamic-templates}.

\begin{algorithm}[h!]
 \KwData{Base corpus $B$, content corpus $C$, unigram model $U$, minimum frequency percentile $p=0.62$, replacement factor $l=25$}
 \KwResult{Text using structure from $b \in B$ and context from $c_1, c_2, ..., c_j \in C$}
 $b \leftarrow selectRandom(B)$ \;
 
 $c_1, \ldots, c_j \leftarrow selectRandom(C, j)$ \;
 
 $M^b \leftarrow createMappingWordsToPOSTags(b)$\;
 
 $M^b \leftarrow filterUnreplaceable(M^b)$\;
 
 $M^b \leftarrow sortAscendingFrequency(M^b, U)$\;
 
 $P^c \leftarrow createMappingPOSTagsToWords(c_1, c_2, \ldots, c_j)$\;
 
 $minR \leftarrow length(b) / l$\;
 
 $minFreq \leftarrow percentile(U, p) $\;
 
 R $\leftarrow$ \{\}\;
 
 \While{$|R| \leq minR$ or ($ |M^b| >0$ and $U(M^b_0) \geq minFreq$) }{
    $w_b \leftarrow M^b_0$;
    
    $M^b \leftarrow M^b \setminus w_b$\;
    
    $candidates \leftarrow P^c(M^b(w_b))$\;
    
    \If{$|candidates| \geq 0$}{
        
       $w_c \leftarrow selectRandom(candidates)$\;
    
   }
    $R \leftarrow R \cup \{(w_b, w_c)\}$\;
    
}
\Return $applyReplacers(b, R)$\;

 \caption{Dynamic Template algorithm in pseudocode}
 \label{alg:dynamic-templates}
\end{algorithm}

% \subsubsection{Example}
The algorithm thus inserts context words into a base text that is used as a template to dynamically respond to the content that needs to be inserted.
This mix of context words into a structure used in a different context can often cause a bisociation, a jump between two frames of references usually present in creative works, between the original narrative and the new context \cite{koestler1964act-of-creation}.
As an example, consider one of the recent tweets by our Twitterbot employing this algorithm, which we translated from Dutch to English. By using an original tweet by Rik Torfs, namely

\begin{center}
$b =$ \textit{``Are there also atheists that don't believe in atheism?''}\footnote{\url{https://twitter.com/torfsrik/status/620144837719932928}}
\end{center}

and combining this with a fragment of a column by the same author as context text

\begin{center}
$c_1, c_2, c_3 $ =
% ``Het feit dat het voormalige Opperwezen dit nieuw verworven inzicht niet probeert te ontkennen, zien zij als een bewijs van hun gelijk. Ook met de Kerk zit het niet snor. Norse pausen.''
\textit{``They see the fact that the former Supreme Being is not trying to deny this newly acquired insight as proof of them being right. Even with the Church, things are not going well. Norse popes.''}\footnote{\url{http://www.standaard.be/cnt/s73709av}}
\end{center}

we get the following result after executing our algorithm, and finding the replacements $R = \{(atheists, popes), (atheism, the\_supreme\_being)\}$\footnote{Note that \textit{``Supreme Being''} in Dutch is only one word (namely \textit{``opperwezen''}), and that \textit{``atheism''} in Dutch has an article, thus providing \textit{``the''}.}:

\begin{center}
$result = $
% Zijn er ook pausen die niet in het opperwezen geloven?
\textit{``Are there also popes that do not believe in the supreme being?''}\footnote{\url{https://twitter.com/TorfsBot/status/1101507600095633410}}
\end{center}

\subsection{Automatic Replying}

There are several methods for making the text generation algorithms interactive, such that the Twitterbot would not only send out stand-alone tweets into the world but also respond to user replies.
One way would be to bias the interpolated Markov model towards words related to words of the conversation by incorporating a factor in $f$ of Equation \ref{eq:weight-influence-markov} that gives a higher weight to relevant context words.
This method could however influence the generator to obsessively use certain rare words, which it would not do in normal circumstances.
Another way of adapting to a user text is inserting relevant context words that have been used by the user as $c$ in our dynamic templates approach.
This is dangerous due to inserting potentially out-of-corpus words from an untrusted source, which attackers might use to make the bot behave offensively and break platform guidelines.

We devised a new method for generating replies using any text generation algorithm without modifying the algorithm.
Since philosophical statements stereotypically tend to be vague and only somewhat related to previous text, the approach first generates philosophical statements like it normally does, and then picks an optimal one to use in the conversation as a reply.
The algorithm first analyses the conversation so far by making a weighted unigram of the words present in it.
The weight of a word in the conversation unigram is multiplied by a factor depending on the author and how recent the text is.
The weight factor is zero for any reply in the conversation longer than ten replies ago, and one if the reply came from the bot.
For the replies of the user, the most recent one has a high weight factor and decreasing linearly for every previous reply, with a minimum total weight factor of one.
The algorithm then generates thousands of random candidates using the interpolated Markov model, as this is a very efficient generator.
It ends by picking the best candidate by looking for the text that has the most rare words in common with weighted unigram coming from the conversation, by summing the weights of the words the text has in common with the conversation and dividing this by the word count in unigram model $U$.
The score of a candidate is also inversely proportional to the difference in length between the last reply and the candidate, such that the bot prefers replying short if the user replies with short answers, and engage using long arguments if the user is also replying using long answers.
This thus helps incorporating topics from earlier in the conversation, and follow the tone of the conversation \footnote{Some (Dutch) examples can be found on \url{https://twitter.com/TorfsBot/status/}$id$ for $id\in \{951851129306009600,  951514464737775616, 937812233085743104\}$}.

\section{Evaluation}

\subsection{Qualitative Performance}

As mentioned earlier, we implemented a Twitterbot called TorfsBot\footnote{\url{https://twitter.com/TorfsBot}}, which imitates tweets made by previous KU Leuven rector Rik Torfs.
This bot uses both the interpolated Markov model and dynamic templates to generate tweets, and uses the interpolated Markov model as text generator for the reply algorithm.
For the interpolated Markov model algorithm, the bot uses both tweets and columns written by Torfs as training data, putting more emphasis on the former by multiplying the weight of the n-grams from this source by ten.
For the dynamic templates, we used the tweets as base texts and the columns for content texts.
TorfsBot has more than 850 followers, making it one of the most popular Dutch Twitterbots\footnote{\url{https://botwiki.org/?s=dutch}}.
We compared the algorithms and their average interactions ($=\frac{likes+retweets}{tweets}$) in Table \ref{tab:algorithm-evaluation-interactions}.
We also compare a subset of the replies, namely those not from a conversation with one single outlier user, who is responsible for 1883 total replies.
From these results, we can see that the locally coherent algorithm (namely the interpolated Markov model) usually produces more prefered content than the globally coherent algorithm (the dynamic template algorithm).
One reason for this result could be that, compared to spotting errors in complicated grammatical structures, it is easier to detect words that can not follow each other, which then drastically lowers the credence in the wisdom of the philosophical statement.

\begin{table}[h]
\caption{Analysing 8582 TorfsBot tweets from September 2016 to April 2019}
\label{tab:algorithm-evaluation-interactions}
\centering
\begin{tabular}{l  r  r}
\toprule
\textit{\textbf{Algorithm}}               & \textit{\textbf{Tweets}} & \textit{\textbf{Avg. Interactions}} \\
\midrule
\textbf{Interpolated Markov Model}        & 2375                                           & 1.11                                                                   \\ 
\textbf{Dynamic Templates}                & 1956                                           & 0.77                                                                    \\
\textbf{Replies}                          & 4271                                           & 0.23                                                                    \\ 
\textbf{Replies without outlier user} & 2388                                           & 0.39                                                                    \\ \bottomrule
\end{tabular}
\end{table}

\subsection{Applicability to Other Domains and Languages}

We also implemented a similar bot using the interpolated Markov models for generating poetry based on the works of Belgian poet Maarten Inghels, called InghelsBot\footnote{\url{https://twitter.com/InghelsBot}}.
This system uses the weight influencers $f$ of Equation \ref{eq:weight-influence-markov} to prefer similar sounding words next to each other using rhyming dictionaries and Levenshtein distance.
It also uses the additional normalisation function $n$ on some of its internal Markov models to normalise to just the part-of-speech tag.
This bot shows that interpolated Markov model works for multiple languages as well as for other text types (namely English and Dutch poetry) without much modifications.

\section{Code}

The code of TorfsBot and all of its dependencies has been made available through \url{https://github.com/twinters/torfs-bot}.
The code to train the interpolated Markov models is also separately available on \url{https://github.com/twinters/markov}

\section{Future Work}

There are several possible improvements to further develop the discussed algorithms.
One large improvement would be using better methods for discovering related words, e.g. by training word embeddings on the training data \cite{mikolov2013word2vec}.
This would allow the dynamic template algorithm to replace the words of the dynamic template with words that have all have similar difference vectors, such that all context words of the resulting sentence are analogous to the context of the base text.
The embeddings could also help the reply generator estimate the relatedness of words better than just plainly using the words themselves.

Another interesting improvement would be interpolating the interpolated Markov model with other sequential generators, such as LSTMs or fine-tuned versions of GPT-2 \cite{radford2019gpt2}, as these are able to keep track of the previously generated texts for much longer and thus produce texts with better grammar.

Since our bots are deployed on Twitter and are thus able to receive constant feedback from users, it would also be interesting to use this feedback mechanism to improve the perceived quality of generated texts.
For example, it might be able to learn which contexts or which base texts work great in dynamic templates.
\section{Conclusion}

We designed and implemented two different algorithms for imitating input text, and made it interactive.
We then showed that the text generated by our systems are well-appreciated by users, and found that the interpolated Markov model got the most positive user feedback, implying that local coherence might be more important than the global structure when generating philosophical statements.

\subsection*{Acknowledgements}

This work was partially supported by Research foundation - Flanders (project G.0428.15).

\bibliographystyle{splncs04}
\bibliography{references}
\end{document}